%% file: main.tex
\documentclass[runningheads]{llncs}
\usepackage{graphicx}
\usepackage{amsmath,amssymb}
\usepackage{float}
\usepackage{graphicx}
\usepackage{algorithm}
\usepackage[noend]{algorithmic}
\usepackage{booktabs}
\usepackage{dblfloatfix}    % To enable figures at the bottom of page
\usepackage[colorlinks = true,
            linkcolor = red,
            urlcolor  = magenta,
            citecolor = red,
            anchorcolor = red,
            backref=page]{hyperref}
\renewcommand{\vec}[1]{\boldsymbol{\mathbf{#1}}}
\begin{document}
\title{Unsupervised semantic discovery \\ through visual patterns detection}
\author{Francesco Pelosin \and
Andrea Gasparetto \and
Andrea Albarelli \and
Andrea Torsello}
\authorrunning{Pelosin et al.}
\institute{Ca' Foscari University, Venice, Italy \\
\email{\{name.surname\}@unive.it}}
\maketitle              % typeset the header of the contribution
\input{sections/abstract_pelosin}
\input{sections/introduction_pelosin}

\input{sections/related_works}

\input{sections/method}
\input{sections/experiments}
\input{sections/conclusions}

\bibliographystyle{splncs04}
\bibliography{biblio}

\end{document}

%% file: sections/abstract_pelosin.tex
\begin{abstract}
    We propose a new fast fully unsupervised method to discover semantic patterns. Our algorithm is able to hierarchically find visual categories and produce a segmentation mask where previous methods fail. Through the modeling of what is a visual pattern in an image, we introduce the notion of ``semantic levels'' and devise a conceptual framework along with measures and a dedicated benchmark dataset for future comparisons. Our algorithm is composed by two phases. A filtering phase, which selects semantical hotsposts by means of an accumulator space, then a clustering phase which propagates the semantic properties of the hotspots on a superpixels basis. We provide both qualitative and quantitative experimental validation, achieving optimal results in terms of robustness to noise and semantic consistency. We also made code and dataset publicly available.

\keywords{visual-pattern-detection  \and semantic-discovery \and cosegmentation}
\end{abstract}

%% file: sections/introduction_pelosin.tex
%%%%%%%%%%%%%%%%%%%%%%%%%%%%%%%%%%%%%%%%%%%%%%%%%%%%%%%%%%%%%%%%%%%%%%%%%%%%%%%%%%5
%                            INTRODUCTION
%%%%%%%%%%%%%%%%%%%%%%%%%%%%%%%%%%%%%%%%%%%%%%%%%%%%%%%%%%%%%%%%%%%%%%%%%%%%%%%%%%5

\section{Introduction}
\label{sec:introduction}

The extraction of semantic categories from images is a fundamental task in image understanding \cite{mottaghi2014role,cordts2016cityscapes,zhou2017scene}. While the task is one that has been widely investigated in the community, most approaches are supervised, making use of labels to detect semantic categories \cite{chen2018encoder}. Comparatively less effort has been put to investigate automatic procedures which enable an intelligent system to learn autonomously extrapolating visual semantic categories without any \textit{a priori} knowledge of the context. 

We observe the fact that in order to define what a visual pattern is, we need to define a scale of analysis (objects, parts of objects etc.). We call these scales \textit{semantic levels} of the real world. Unfortunately most influential models arising from deep learning approaches still show a limited ability over scale invariance \cite{singh2018analysis,li2019scale} which instead is common in nature. In fact, we don't really care much about scale, orientation or partial observability in the semantic world. For us, it is way more important to preserve an ``internal representation'' that matches reality \cite{DiCarlo2012HowDT,logothetis1996visual}.

\begin{figure}[t]
  \centering
  \includegraphics[width=1\linewidth]{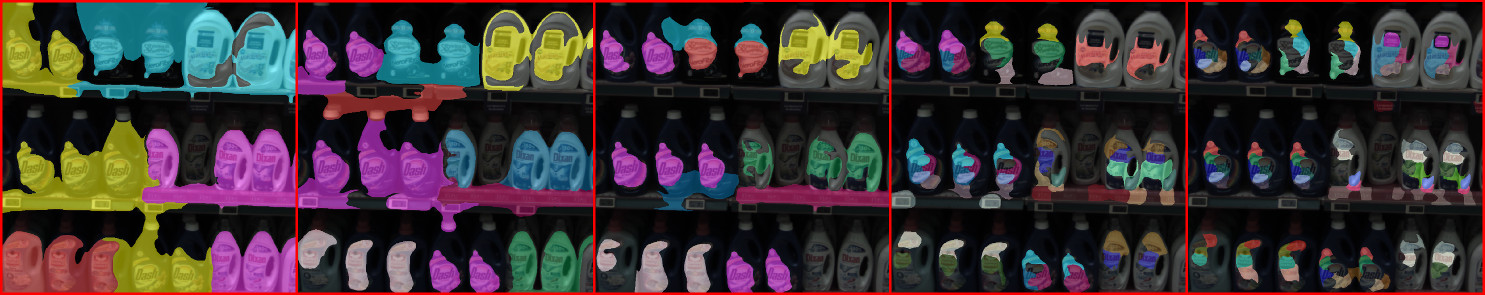}
  \caption{A real world example of unsupervised segmentation of a grocery shelf. Our method can automatically discover both low-level coherent patterns (brands and logos) and high-level compound objects (detergents) by controlling the semantic level of the detection and segmentation process.}
  \label{fig:levels}
\end{figure}

Our method leverages repetitions (Figure \ref{fig:levels}) to capture the internal representation in the real world and then extrapolates categories at a specific semantic level. We do this without continuous geometrical constraints on the visual pattern disposition, which is common among other methodologies \cite{DBLP:conf/cvpr/PrittsCM14,DBLP:conf/cvpr/PrittsCM14,DBLP:conf/wacv/LettryPVG17,DBLP:journals/cg/Rodriguez-Pardo19,DBLP:conf/cvpr/HeZRS16}. 

We also do not constrain ourselves to find only one visual pattern, which is another very common assumption. Indeed, what if the image has more than one visual pattern? One can observe that this is \textit{always} the case. Each visual repetition can be hierarchically decomposed in its smaller parts which, in turn, repeat over different semantic levels. This peculiar observation allow our work to contribute to the community as follows:

\begin{itemize}
    \item A new pipeline able to capture semantic categories with the ability to hierarchically span over semantic levels.
    \item A better conceptual framework to evaluate analogous works through the introduction of the semantic levels notion along with a new metric.
    \item A new benchmark dataset of 208 labelled images for visual repetition detection. 
\end{itemize}

Code, dataset and notebooks are public and available at:  \texttt{\url{https://git.io/JT6UZ}}.

%% file: sections/related_works.tex
%%%%%%%%%%%%%%%%%%%%%%%%%%%%%%%%%%%%%%%%%%%%%%%%%%%%%%%%%%%%%%%%%%%%%%%%%%%%%%%%%%5
%                            RELATED WORKS
%%%%%%%%%%%%%%%%%%%%%%%%%%%%%%%%%%%%%%%%%%%%%%%%%%%%%%%%%%%%%%%%%%%%%%%%%%%%%%%%%%5

\section{Related Works} 
\label{sec:relatedworks}

Several works have been proposed to tackle visual pattern discovery and detection. While the paper by Leung and Malik \cite{DBLP:conf/eccv/LeungM96} could be considered seminal, many other works build on their basic approach, working by detecting contiguous structures of similar patches by knowing the window size enclosing the distinctive pattern. 

One common procedure in order to describe what a pattern is, consists to first extract descriptive features such as SIFT to perform a clustering in the feature space and then model the group disposition over the image by exploiting geometrical constraints, as in \cite{DBLP:conf/cvpr/PrittsCM14} and \cite{DBLP:conf/cvpr/ChumM10}, or by relying only on appearance, as in \cite{DBLP:conf/icpr/DoubekMPC10,DBLP:conf/cvpr/LiuL13,DBLP:journals/pami/ToriiSOP15}.

The geometrical modeling of the repetitions usually is done by fitting a planar 2-D lattice, or a deformation of it \cite{DBLP:journals/pami/ParkBCL09}, through RANSAC procedures as in \cite{DBLP:conf/bmvc/SchaffalitzkyZ98} \cite{DBLP:conf/cvpr/PrittsCM14} or even by exploiting the mathematical theory of crystallographic groups as in \cite{DBLP:journals/pami/LiuCT03}. Shechtman and Irani  \cite{DBLP:conf/cvpr/ShechtmanI07}, also exploited an active learning environment to detect visual patterns in a semi-supervised fashion. For example Cheng et al. \cite{DBLP:journals/tog/ChengZMHH10} use input scribbles performed by a human to guide detection and extraction of such repeated elements, while Huberman  and Fattal \cite{DBLP:conf/cvpr/HubermanF16} ask the user to detect an object instance and then the detection is performed by exploiting correlation of patches near the input area.

Recently, as a result of the new wave of AI-driven Computer Vision, a number of Deep Leaning based approaches emerged, in particular Lettry et al. \cite{DBLP:conf/wacv/LettryPVG17} argued that filter activation in a model such as AlexNet can be exploited in order to find regions of repeated elements over the image, thanks to the fact that filters over different layers show regularity in the activations when convolved with the repeated elements of the image. On top of the latter work, Rodríguez-Pardo et al. \cite{DBLP:journals/cg/Rodriguez-Pardo19} proposed a modification to perform the texture synthesis step. 

A brief survey of visual pattern discovery in both video and image data, up to 2013, is given by Wang et al. \cite{DBLP:journals/widm/WangZY14}, unfortunately after that it seems that the computer vision community lost interest in this challenging problem. We point out that all the aforementioned methods look for \textit{only one} particular visual repetition except for \cite{DBLP:conf/cvpr/LiuL13} that can be considered the most direct competitor and the main benchmark against which to compare our results.

%% file: sections/method.tex
\section{Method Description}
\label{sec:methoddescription}

%%%%%%%%%%%%%%%%%%%%%%%%%%%%%%%%%%%%%%%%%%%%%%%%%%%%%%%%%%%%%%%%%%%%%%%%%%%%%%%%%%5
\subsection{Features Localization and Extraction}
\label{sec:featureslocalizationextraction}
%%%%%%%%%%%%%%%%%%%%%%%%%%%%%%%%%%%%%%%%%%%%%%%%%%%%%%%%%%%%%%%%%%%%%%%%%%%%%%%%%%5

% Consider an RGB image $\mathcal{I} \in [0,1]^{n\times m \times 3}$,
We observe that any visual pattern is delimited by its contours. The first step of our algorithm, in fact, consists in the extraction of a set $\mathcal{C}$ of contour \emph{keypoints} indicating a position $\vec{c}_{j}$ in the image. To extract keypoints, we opted for the Canny algorithm, for its simplicity and efficiency, although more recent and better edge extractor could be used \cite{liu2019richer} to have a better overall procedure.

A descriptor $d_{j}$ is then computed for each selected $\vec{c}_{j} \in \mathcal{C}$ thus obtaining a \emph{descriptor set} $\mathcal{D}$. In particular, we adopted the DAISY algorithm because of its appealing dense matching properties that nicely fit our scenario. Again, here we can replace this module of the pipeline with something more advanced such as \cite{DBLP:conf/nips/OnoTFY18} at the cost of some computational time.

%%%%%%%%%%%%%%%%%%%%%%%%%%%%%%%%%%%%%%%%%%%%%%%%%%%%%%%%%%%%%%%%%%%%%%%%%%%%%%%%%%5
\subsection{Semantic Hot Spots Detection}
\label{sec:semantichotspotdetection}
%%%%%%%%%%%%%%%%%%%%%%%%%%%%%%%%%%%%%%%%%%%%%%%%%%%%%%%%%%%%%%%%%%%%%%%%%%%%%%%%%%5

\begin{figure*}[t!]
  \centering
  \includegraphics[width=1\linewidth]{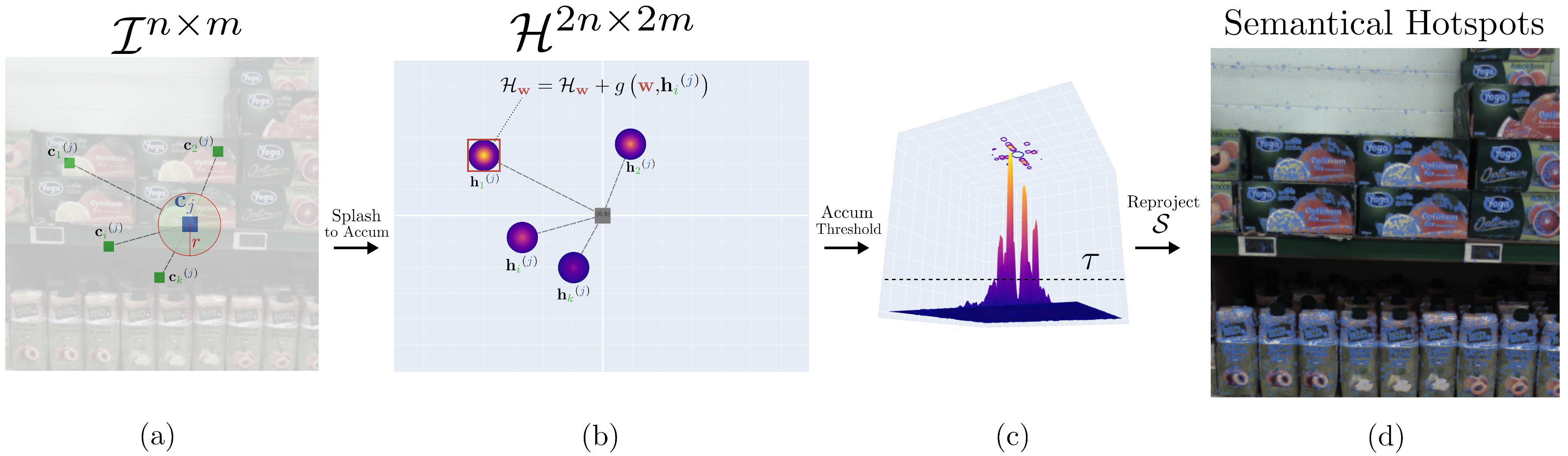}
  \caption{(a) A splash in the image space with center in the keypoint $\vec{c}_j$. (b) $\mathcal{H}$, with the superimposed splash at the center, you can note the different levels of the vote ordered by endpoint importance i.e. descriptor similarity. (c) 3D projection showing the gaussian-like formations and the thresholding procedure of $\mathcal{H}$. (d) Backprojection through the set $\mathcal{S}$.}
  \label{fig:houghaccumulator}
\end{figure*}

In order to detect self-similar patterns in the image we start by associating the $k$ most similar descriptors for each descriptor $\vec{d}_j$. We can visualize this data structure as a star subgraph with $k$ endpoints called \emph{splash} ``centered'' on descriptor $\vec{d}_{j}$. Figure \ref{fig:houghaccumulator} (a) shows one.

Splashes potentially encode repeated patterns in the image and similar patterns are then represented by similar splashes.
The next step consists in separating these splashes from those that encode noise only, this is accomplished through an accumulator space.

In particular, we consider a $2$-D \emph{accumulator space} $\mathcal{H}$ of size double the image. We then superimpose each splash on the space $\mathcal{H}$ and cast $k$ votes as shown in Figure \ref{fig:houghaccumulator} (b). In order to take into account the noise present in the splashes, we adopt a gaussian vote-casting procedure $g(\cdot)$. Similar superimposed splashes contribute to similar locations on the accumulator space, resulting in peak formations (Figure \ref{fig:houghaccumulator} (c)). We summarize the voting procedure as follows:

\begin{equation}
    \mathcal{H}_{\vec{w}} = \mathcal{H}_{\vec{w}} + g(\vec{w}, \vec{h}^{(j)}_{i})
\end{equation}
where $\vec{h}^{(j)}_{i}$ is the $i$-th splash endpoint of descriptor $\vec{d}_j$ in accumulator coordinates and $\vec{w}$ is the size of the gaussian vote. We filter all the regions in $\mathcal{H}$ which are above a certain \emph{threshold} $\tau$, to get a set $\mathcal{S}$ of the locations corresponding to the peaks in $\mathcal{H}$. The $\tau$ parameter acts as a coarse filter and is not a critical parameter to the overall pipeline. A sufficient value is to set it to $0.05 \cdot max(\mathcal{H})$. 
Lastly, in order to visualize the semantic hotspots in the image plane we map splash locations between $\mathcal{H}$ and the image plane by means of a \emph{backtracking structure} $\mathcal{V}$.

In summary, the key insight here is that similar visual regions share similar splashes, we discern noisy splashes from representative splashes through an auxiliary structure, namely an accumulator. We then identify and backtrack in the image plane the semantic hotspots that are candidate points part of a visual repetition.

%%%%%%%%%%%%%%%%%%%%%%%%%%%%%%%%%%%%%%%%%%%%%%%%%%%%%%%%%%%%%%%%%%%%%%%%%%%%%%%%%%5
\subsection{Semantic Categories Definition and Extraction}
\label{sec:graph}
%%%%%%%%%%%%%%%%%%%%%%%%%%%%%%%%%%%%%%%%%%%%%%%%%%%%%%%%%%%%%%%%%%%%%%%%%%%%%%%%%%5

While the first part previously described acts as a filter for noisy keypoints allowing to obtain a good pool of candidates, we now transform the problem of finding visual categories in a problem of dense subgraphs extraction.

We enclose semantic hotspots in superpixels, this extends the semantic significance of such identified points to a broader, but coherent, area. To do so we use the SLIC \cite{DBLP:journals/pami/AchantaSSLFS12} algorithm which is a simple and one of the fastest approaches to extract superpixels as pointed out in this recent survey \cite{DBLP:journals/cviu/StutzHL18}. Then we choose the cardinality of the \emph{superpixels} $\mathcal{P}$ to extract. This is the second and most fundamental parameter that will allow us to span over different semantic levels.

Once the superpixels have been extracted, let $\mathcal{G}$ be an \emph{undirected weighted graph} where each node correspond to a superpixel $p \in \mathcal{P}$. In order to put edges between graph nodes (i.e. two superpixels), we exploit the splashes origin and endpoints. In particular the strength of the connection between two vertices in $\mathcal{G}$ is calculated with the number of splashes endpoints falling between the two in a mutual coherent way. So to put a weight of 1 between two nodes we need exactly 2 splashes endpoints falling with both origin and end point in the two candidate superpixels.

With this construction scheme, the graph has clear dense subraphs formations. Therefore, the last part simply computes a partition of $\mathcal{G}$ where each connected component correspond to a cluster of similar superpixels. In order to achieve such objective we optimize a function that is maximized when we partition the graph to represent so. To this end we define the following \textit{density score} that given $G$ and a set $K$ of connected components captures the optimality of the clustering:

\begin{equation}
    s(G, K) = \sum_{k \in K} \mu(k) - \alpha \left | K \right |
\end{equation}
where $\mu(k)$ is a function that computes the average edge weight in a undirected weighted graph.

\begin{algorithm}[t]
    \caption{Semantic categories extraction algorithm}
    \vspace{2mm}
\begin{algorithmic}
\REQUIRE $G$ weighted undirected graph
\STATE $i=0$
\STATE $s^{*}=-\inf$
\STATE $K^{*}= \emptyset$
\WHILE{$G_{i}$ is not fully disconnected}
    \STATE $i = i + 1$
    \STATE Compute $G_{i}$ by corroding each edge with the minimum edge weight
    \STATE Extract the set $K_{i}$ of all connected components in $G_{i}$
    \STATE $s(G_{i}, K_{i}) = \sum_{k \in K_{i}} \mu(k) - \alpha \left | K_{i} \right |$
    \IF{$s(G_{i}, K_{i}) > s^{*}$}
    \STATE $s^{*} = s(G_{i}, K_{i})$
    \STATE $K^{*} = K_{i}$
    \ENDIF
\ENDWHILE
\RETURN $s^{*}, K^{*}$ 
\end{algorithmic}
\label{alg:graphcorrosion}
\end{algorithm}

The first term, in the score function, assign a high vote if each connected component is dense. While the second term acts as a regulator for the number of connected components. We also added a weighting factor $\alpha$ to better adjust the procedure. As a proxy to maximize this function we devised an \emph{iterative algorithm} reported in Algorithm \ref{alg:graphcorrosion} based on graph corrosion and with temporal complexity of $O(\left | E \right |^{2} + \left | E \right | \left | V \right |)$. At each step the procedure corrupts the graph edges by the minimum edge weight of $G$. For each corroded version of the graph that we call \emph{partition}, we compute $s$ to capture the density. Finally the algorithm selects the corroded graph partition which maximizes the $s$ and subsequently extracts the node groups.

In brevity we first enclose semantic hotspots in superpixels and consider each one as a node of a weighted graph. We then put edges with weight proportional to the number of splashes falling between two superpixels. This results in a graph with clear dense subgraphs formations that correspond to superpixels clusters i.e. \textit{semantic categories}. The semantic categories detection translates in the extraction of dense subgraphs. To this end we devised an iterative algorithm based on graph corrosion where we let the procedure select the corroded graph partition that filters noisy edges and let dense subgraphs emerge. We do so by maximizing score that captures the density of each connected component.

%% file: sections/experiments.tex
%%%%%%%%%%%%%%%%%%%%%%%%%%%%%%%%%
\section{Experiments}
\label{sec:experimental}
%%%%%%%%%%%%%%%%%%%%%%%%%%%%%%%%%

\begin{figure*}[t!]
  \centering
  \includegraphics[width=\linewidth]{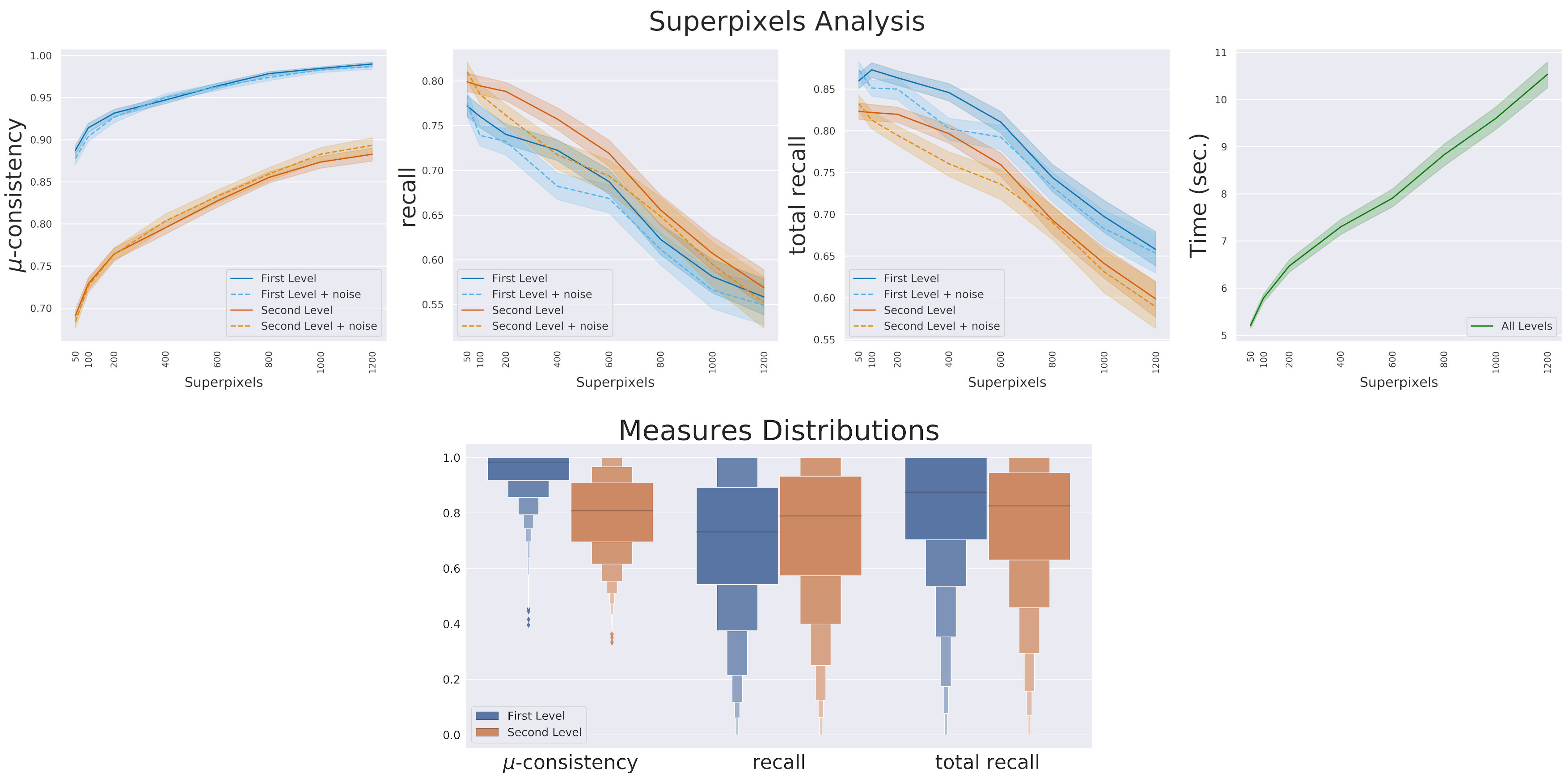}
  \caption{(top) Analysis of measures as the number of superpixels $|\mathcal{P}|$ retrieved varies. The rightmost figure shows the running time of the algorithm.  We repeated the experiments with the noisy version of the dataset but report only the mean since variation is almost equal to the original one. (bottom) Distributions of the measures for the two semantic levels, by varying the two main parameters $r$ and $|\mathcal{P}|$.}
  \label{fig:superpixels}
\end{figure*}

\begin{figure}[t]
  \centering
  \includegraphics[width=1\linewidth]{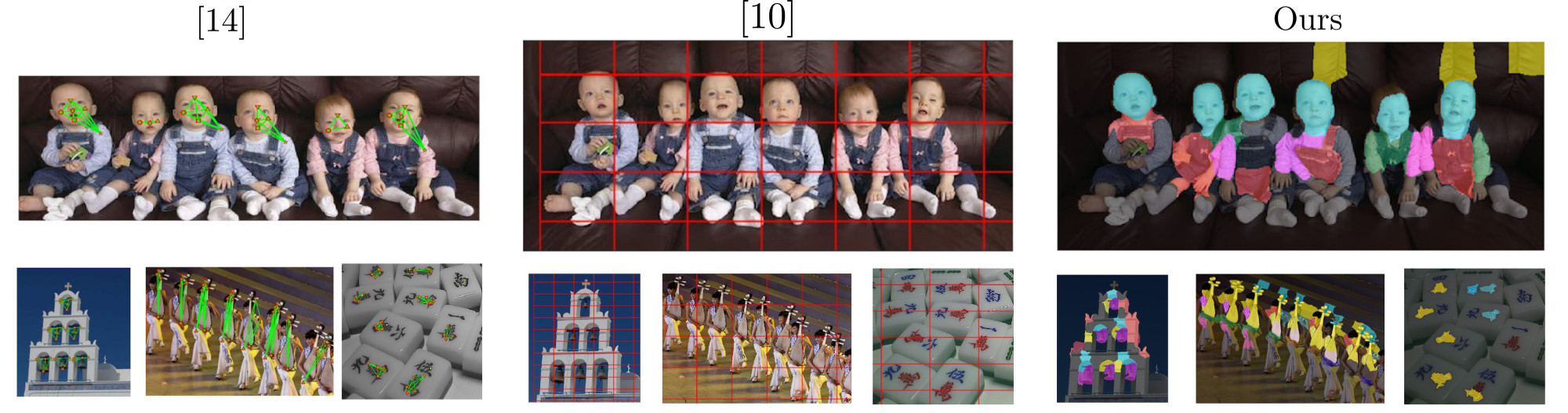}
  \caption{Qualitative comparison between \cite{DBLP:conf/cvpr/LiuL13}, \cite{DBLP:conf/wacv/LettryPVG17} and our algorithm. Our method detects and segments more than one pattern and does not constrain itself to a particular geometrical disposition.}
  \label{fig:distributions}
\end{figure}

\subsubsection{Dataset}
 As we introduced in Section \ref{sec:introduction} one of the aims of this work is to provide a better comparative framework for visual pattern detection. To do so we created a public dataset by taking 104 pictures of store shelves. Each picture has been took with a 5mpx camera with approximatively the same visual conditions. We also rectified the images to eliminate visual distortions. 

 We manually segmented and labeled each repeating product in two different semantic levels. In the \textbf{first semantic level} \emph{products made by the same company} share the same label. In the \textbf{second semantic level} visual repetitions consist in the \emph{exact identical products}. In total the dataset is composed by 208 ground truth images, half in the first level and the rest for the second one.

\subsubsection{$\mu$-consistency}
We devised a new measure that captures the semantic consistency of a detected pattern that is a proxy of the average precision of detection.

In fact, we want to be sure that all pattern instances fall on similar ground truth objects. First we introduce the concept of semantic consistency for a particular pattern $\vec{p}$. Let $\vec{P}$ be the set of patterns discovered by the algorithm. Each pattern $\vec{p}$ contains several instances $\vec{p}_{i}$. $\vec{L}$ is the set of ground truth categories, each ground truth category $\vec{l}$ contain several objects instances $\vec{l}_{i}$. Let us define $\vec{t}_{p}$ as the vector of ground truth labels touched by all instances of $\vec{p}$. We say that $\vec{p}$ is consistent if all its instances $\vec{p}_{i}, i=0\dots |\vec{p}|$ fall on ground truth regions sharing the same label. In this case $\vec{t}_{p}$ would be uniform and we consider $\vec{p}$ a good detection. The worst scenario is when given a pattern $\vec{p}$ every $\vec{p}_{i}$ falls on objects with different label $\vec{l}$ i.e. all the values in $\vec{t}_{p}$ are different. 

To get an estimate of the overall consistency of the proposed detection, we average the consistency for each $\vec{p} \in \vec{P}$ giving us: 

\begin{equation}
    \text{$\mu$-consistency} =  \frac{1}{\left | \vec{P} \right |} \sum_{\vec{p} \in \vec{P}} \frac{\left| \operatorname{mode}\left(\vec{t}_{p}\right)\right|}{\left|\vec{t}_{p}\right|}
\end{equation}

\subsubsection{recall}
The second measure is the classical recall over the objects retrieved by the algorithm. Since our object detector outputs more than one pattern we average the recall for each ground truth label by taking the best fitting pattern.

\begin{equation}
    \frac{1}{\left | \vec{L} \right |} \sum_{\vec{l} \in \vec{L}} \operatorname{max}_{\vec{p} \in \vec{P}} \text { recall }(\vec{p}, \vec{l})
\end{equation}

The last measure is the \textbf{total recall}, here we consider a hit if any of the pattern falls in a labeled region. In general we expect this to be higher than the recall.

We report the summary performances in Figure \ref{fig:distributions}. As can be seen the algorithm achieves a very high $\mu$-consistency while still able to retrieve the majority of the ground truth patterns in both levels.

One can observe in Figure \ref{fig:superpixels} an inverse behaviour between recall and consistency as the number of superpixels retrieved grows. This is expected since less superpixels means bigger patterns, therefore it is more likely to retrieve more ground truth patterns. 

In order to study the robustness we repeated the same experiments  with an altered version of our dataset. In particular for each image we applied one of the following corruptions: Additive Gaussian Noise ($scale=0.1*255$), Gaussian Blur  ($\sigma = 3$), Spline Distortions (grid affine), Brightness ($+100$), and Linear Contrast ($1.5$).

\subsubsection{Qualitative Validation}
Firstly we begin the comparison by commenting on \cite{DBLP:conf/cvpr/LiuL13}. One can observe that our approach has a significant advantage in terms of how the visual pattern is modeled. While the authors model visual repetitions as geometrical artifacts associating points, we output a higher order representation of the visual pattern. Indeed the capability to provide a segmentation mask of the repeated instance region together the ability to span over different levels unlocks a wider range of use cases and applications.

As qualitative comparison we also added the latest (and only) deep learning based methodology \cite{DBLP:conf/wacv/LettryPVG17} we found. This methodology is only able to find a single instance of visual pattern, namely the most frequent and most significant with respect to the filters weights. This means that the detection strongly depends from the training set of the CNN backbone, while our algorithm is fully unsupervised and data agnostic. 

\subsubsection{Quantitative Validation}
We compared quantitatively our method against \cite{DBLP:conf/cvpr/LiuL13} that constitutes, to the best of our knowledge, the only work developed able to detect more than one visual pattern. We recreated the experimental settings of the authors by using the Face dataset \cite{DBLP:journals/cviu/Fei-FeiFP07} as benchmark achieving $1.00$ precision vs. $0.98$ of \cite{DBLP:conf/cvpr/LiuL13} and $0.77$ in recall vs. and $0.63$. We considered a miss on the object retrieval task, if more than 20\% of a pattern total area falls outside from the ground truth. The parameter used were $|\mathcal{C}|=9000$, $k=15$, $r=30$, $\tau=5$, $| \mathcal{P} |=150$. We also fixed the window of the gaussian vote to be $11 \times11$ pixels throughout all the experiments.

%% Comparison table
%\begin{table}[t!]
%\centering
%\normalsize
%\begin{tabular}{@{}lllll@{}}
%\toprule
%\multicolumn{5}{c}{Face Dataset %\cite{DBLP:journals/cviu/Fei-FeiFP07}}                                                    \\ \midrule
 %                & \multicolumn{2}{l}{Precision} & \multicolumn{2}{l}{Recall}   \\
%GRASP \cite{DBLP:conf/cvpr/LiuL13}   & \multicolumn{2}{l}{0.98}    & \multicolumn{2}{l}{0.63} \\ 
%\textbf{Ours}      & \multicolumn{2}{l}{\textbf{1.00}}    & \multicolumn{2}{l}{\textbf{0.77}}  \\ \bottomrule
%\\
%\end{tabular}
%\caption{\label{tab:grasp}Object-Level precision/recall rates. Parameters used: $|\mathcal{C}|=9000$, $k=15$, $r=30$, $\tau=5$, $| \mathcal{P} |=150$}
%\vspace{-1mm}
%\end{table}

%\begin{table}[h]
%\centering
%\normalsize
%\begin{tabular}{@{}lllll@{}}
%\toprule
%\multicolumn{5}{c}{Face Dataset \cite{DBLP:journals/cviu/Fei-FeiFP07}}                                                    \\ \midrule
                 %& \multicolumn{2}{l}{\cite{DBLP:conf/cvpr/LiuL13}} & %\multicolumn{2}{l}{\textbf{Ours}}   \\
%Precision (\%)   & \multicolumn{2}{l}{0.98}    & \multicolumn{2}{l}{\textbf{1.00}} \\ 
%Recall (\%)      & \multicolumn{2}{l}{0.63}    & \multicolumn{2}{l}{\textbf{0.77}}  \\ %\bottomrule
%\\
%\end{tabular}
%\caption{\label{tab:grasp}Object-Level precision/recall rates. Parameters used: $|\mathcal{C}|=9000$, $k=15$, $r=30$, $\tau=5$, $| \mathcal{P} |=150$}
%\vspace{-1mm}
%\end{table}

%% file: sections/conclusions.tex
%%%%%%%%%%%%%%%%%%%%%
\section{Conclusions}
\label{sec:conclusions}
%%%%%%%%%%%%%%%%%%%%%

With this paper we introduced a fast and unsupervised method addressing the problem of finding semantic categories by detecting consistent visual pattern repetitions at a given scale. The proposed pipeline hierarchically detects self-similar regions represented by a segmentation mask.

As we demonstrated in the experimental evaluation, our approach retrieves more than one pattern and achieves better performances with respect to competitors methods. We also introduce the concept of \emph{semantic levels} endowed with a dedicated dataset and a new metric to provide to other researchers tools to evaluate the consistency of their approaches.

\subsection{Acknowledgments}
We would like to express our gratitude to Alessandro Torcinovich and Filippo Bergamasco for their suggestions to improve the work. We also thank Mattia Mantoan for his work to produce the dataset labeling.